# *Scalable Modular Synthetic Data Generation for Advancing Aerial Autonomy*

Mehrnaz Sabet [a,*], Praveen Palanisamy [b], Sakshi Mishra [b]

[a] Department of Information Science, Cornell University, Ithaca, NY, USA
[b] Microsoft, Bellevue, WA, USA

ms3662@cornell.edu, praveen.palanisamy@microsoft.com, sakshimishra@microsoft.com

## Abstract

One major barrier to advancing aerial autonomy has been collecting large-scale aerial datasets for training machine learning models. Due to costly and time-consuming real-world data collection through deploying drones, there has been an increasing shift towards using synthetic data for training models in drone applications. However, to increase widespread generalization and transferring models to real-world, increasing the diversity of simulation environments to train a model over all the varieties and augmenting the training data, has been proved to be essential. Current synthetic aerial data generation tools either lack data augmentation or rely heavily on manual workload or real samples for configuring and generating diverse realistic simulation scenes for data collection. These dependencies limit scalability of the data generation workflow. Accordingly, there is a major challenge in balancing generalizability and scalability in synthetic data generation. To address these gaps, we introduce a scalable Aerial Synthetic Data Augmentation (ASDA) framework tailored to aerial autonomy applications. ASDA extends a central data collection engine with two scriptable pipelines that automatically perform scene and data augmentations to generate diverse aerial datasets for different training tasks. ASDA improves data generation workflow efficiency by providing a unified prompt-based interface over integrated pipelines for flexible control. The procedural generative approach of our data augmentation is performant and adaptable to different simulation environments, training tasks and data collection needs. We demonstrate the effectiveness of our method in automatically generating diverse datasets and show its potential for downstream performance optimization. Our work contributes to generating enhanced benchmark datasets for training models that can generalize better to real-world situations.

**Video:** **youtube.com/watch?v=eKpOh-K-NfQ**

*Keywords:* aerial autonomy, drone, synthetic data, sim-to-real, domain randomization

## 1 INTRODUCTION

Drones are poised to change the traditional mediums of delivery, inspection, urban mobility, farming, and wildfire monitoring for the better. Despite the many benefits that drones offer towards a sustainable future, integrating them in different applications hasn't made much progress during the last few years. Accelerating the integration process requires reliable aerial autonomy which has been identified as the main challenge for slow progress and regulatory challenges [1, 2, 3, 4]. To advance autonomy in vision-based navigation and detection, Machine Learning (ML) models require a vast amount of aerial datasets for training but data collection through deploying drones and testing models and algorithms in different trials is time-consuming, costly and limited [5, 6, 7]. Accordingly, synthetic datasets are increasingly being used for training ML models in a range of vision-based tasks [6, 7, 8, 9, 10, 11, 12].

---

[*] Corresponding author.

But current aerial synthetic data generation methods are based on limited training simulation scenes that can cause the model to overfit and as a result, generalize poorly to unseen environments [7]. Lack of diversity in the simulation scenes can also increase the reality gap between the synthetic and the real world (sim-to-real) that causes performance issues when transferring models due to distribution difference [7, 13]. To increase generalizability of trained models and policies to different domains and situations of the real-world, models need to be trained over a diverse set of datasets through different variations of the simulation environment. Augmenting the data through Domain Randomization (DR) namely increasing the diversity of simulation environments to train a model over all the varieties, has been proved to be very effective in addressing this problem [7, 13, 14, 15, 16, 17, 18]. Current synthetic data generation methods either lack incorporated data augmentation workflows or have increasing dependency on manual workload for configuring simulation parameters and designing simulation scenes [5, 7, 20, 22] which becomes error-prone, and time-consuming for large-scale diverse scene generation. Additionally, some methods rely on real samples [15, 18, 21, 22, 23] to inform and increase scene diversity which limits the scalability of data augmentation. Scalability of the synthetic data augmentation method is important so that we can generate tens of thousands of environment variations that can be easily used and adapted for data generation in different training tasks without significant workload, human intervention or change in the process. On the other hand, DR workflows adapted for self-driving car scenarios [14, 15, 19, 20, 21] cannot be transferred to aerial data generation because the environment requirements and domain gap for aerial imageries significantly differs.

By focusing on the need for rapid generalization in aerial synthetic data generation workflows to solve reviewed problems, we present a new procedural generative Aerial Synthetic Data Augmentation (ASDA) framework that extends a central data collection engine by adding two pre- and post- processing scriptable pipelines that automatically generate diverse simulation scenes and augment the generated data. The pre-processing pipeline has built-in randomization operations that can be combined and applied automatically to different scene specifications that seed its generative scene augmentation process. The pipeline is implemented such that it can run a batch of operations attached to it to generate unbounded number of scene variations based on operation specifications to augment simulation scenes for data collection. The pipeline allows classification and customization of operations based on the training task type, environment type, or other data collection needs. This design increases flexibility and scalability of processing operations and contributes to performance optimization for parameter mapping based on downstream tasks. The post-processing pipeline has built-in augmentation operations that are applied over collected data to further augment the training datasets. Additionally, ASDA increases synthetic data generation efficiency by connecting pre- and post- pipelines to the data collection engine and enabling flexible control through a unified prompt-based interface. The two pre- and post- processing pipelines have 50 built-in scene and data augmentation operations in total and can be further extended to include more customized operations for added functionality.

ASDA's generative scene augmentation is enabled through a layered DR approach incorporated to its pre-processing pipeline with an extensive parameter set to support any aerial vision-based detection or navigation training task. Through the layered DR, randomization operations are separated into multiple layers that are incrementally applied on top of one another when automatically generating variations of the simulation scenes in the pre-processing pipeline. The operations are applied in ordered layers of: asset variation generation, asset distribution, obstacle generation, global variation generation and drone trajectory pose generation. The main challenge for generating realistic variations of the scenes is to distribute generated variations of each asset in valid locations to avoid unrealistic scenes that can hinder the trained policy's performance in a negative way. In our approach, we embed 3D scene graph information of the environment into distribution operations in pre-processing pipeline to automatically create a distribution space for asset objects over base maps of environments. Accordingly, randomized variations of each type of asset are distributed over valid location spaces mapped as a sub-graph on a separate distribution layer from the scene. Utilizing scene graph to extract a distribution layer per asset object together with operations of next layers to incrementally apply randomizations over the environment makes our DR method adaptive to any type of environment. Accordingly, ASDA's workflow is agnostic to the underlying environment type meaning that it can generate new variations of simulation scenes from any type of base map, whether a carefully curated 3D scene, a 3D map or a Block-NeRF [24], without needing to change the data generation procedure.

We demonstrate the effectiveness of our data augmentation framework in automatically generating a diverse dataset by running an experiment for a drone landing pad detection task. **Our contributions** are: 1) ASDA, new procedural generative data augmentation framework that allows for the performant generation of an unbounded number of diverse realistic simulation scenes for aerial synthetic data collection. Our approach results in 75% more variants in the end data and increases data generation efficiency by 50% compared to manual procedural workflows of previous methods. 2) A novel multi-layer adaptive domain randomization approach guided by scene graph information to automatically distribute objects in valid spatial arrangements over simulation scenes allowing adaption of ASDA's generative scene augmentation to any type of base environment  3) Increasing aerial synthetic data generation efficiency by connecting scene and data augmentation pipelines to data collection engine and enabling flexible control through a unified prompt-based interface which is important as many automatic methods provide little to no control 4) Proposing a new iterative data collection optimization method based on ASDA's workflow by learning best operation policies to use to meet a performance target.

We hope that the new data generation architecture further contributes to advancing aerial autonomy by accelerating the development of large-scale diverse benchmark datasets and pre-trained models that can widely adapt to new environments and transfer better to the real-world.

## 2    RELATED WORK

Using synthetic datasets for training Machine Learning (ML) models has been increasingly used in a range of vision-based tasks where labeling images is tedious and cost of real-world data collection is high [9, 10, 11, 12]. Similarly, limited aerial datasets have resulted in utilizing synthetic data for training ML models in drone-based applications [6, 7, 8, 25]. The challenges of manual data collection through drones are associated with their limited batteries and payloads, and different, expensive sensor setup contributing to costly, time-consuming data collection [18].

### 2.1   Synthetic aerial data generation

Using simulation platforms to collect synthetic data for training machine learning models has been on the rise in recent years. While most of these simulation platforms are targeted at self-driving cars, only a few advanced simulation software has been developed for synthetic aerial data collection. The two dominant simulation platforms for synthetic aerial data collection in research community, are Flightmare [26] and AirSim [6]. Both of these simulators are model-driven meaning that they rely on predefined models of scenery and underlying physics with a focus on high quality visual appearances for collecting synthetic aerial data through simulated drone agents. Compared to Flightmare that mostly focuses on providing a modular physics engine, AirSim provides a configurable high-fidelity simulation for training ML models by facilitating multi-modal aerial synthetic data collection. AirSim has been dominantly used in relevant robotics and computer vision research for advancing ML models for aerial autonomy [7, 16]. While AirSim supports different types of pre-defined simulation environments such as 3D maps or pre-designed simulation scenes for data collection, it follows a procedural model for configuration of simulation scenes. Procedural model relies heavily on manual workload for tunning simulation scene parameters and placing objects in the scene which limits the ability to automatically generate diverse augmentations of the simulation scenes before data collection. Accordingly generated data is limited in diversity which can hinder generalizability of trained models to new environments.

Another recent scalable data generation method, TOPO-DataGen [27] generates large-scale multi-modal aerial synthetic data by using off-the-shelf geodata as input to a rendering engine. While the method is able to generate large-scale datasets for aerial localization tasks, it lacks scene diversity to increase generalizability of trained models thus the model is prone to failure when encountered with situations not seen in the training data which is also mentioned as a limitation in the paper. Additionally, the method lacks flexibility in configuring simulation environments for other types of training tasks compared to AirSim which is why we chose AirSim as the central data collection engine for ASDA's augmentation pipelines.

## 2.2 Sim-to-real gap

While synthetic datasets are a promising alternative to resolve data insufficiency for training tasks, the domain gap between the synthetic and the real-world (sim-to-real) often raises performance issues when transferring models [13, 18, 22]. The sim-to-real gap has been associated with two types of distribution differences between the domains: appearance gap and content gap [15, 22]. While the appearance gap refers to the difference in appearance distribution between the synthetic images and the real samples, the content gap suggests that the problem is also associated with the lack of diversity and distribution of objects (such as type and spatial arrangement of objects) in the synthetic data that is not representative of the diverse situations in real-world [15]. Accordingly, a model trained on a dataset with a specific lighting, location, or weather does not transfer well to a different real-world lighting, location, or weather situation. A large body of work has been focused on closing the appearance gap by updating the distributions in synthetic images to look closer to the real samples through a mapping enforced by a task model often built on generative adversarial networks (GAN) [28]. On the other hand, less work has looked into addressing the content gap in the data generation workflow which is more challenging [15]. Our goal in implementing ASDA is to address the content gap by generating diverse datasets through scene augmentations for different training tasks.

## 2.3 Domain Randomization

Proposed approaches for increasing generalizability of trained models over synthetic datasets can be mainly categorized into Domain Adaption (DA) and Domain Randomization (DR). Synthetic data generation methods based on DA use a data-driven approach to learn how to generate realistic scenes representative of situations seen in a set of ground truth real samples [14, 15, 23]. Due to dependency on availability of a decent amount of real data samples, DA methods have limited scalability especially for drone-based applications as the real-samples are very scarce [7]. Compared to DA, DR offers a more scalable approach to addressing the content gap as it's not dependent on real samples. DR suggests generating a variety of simulation environments with randomized parameters and training the models across all of them to increase the generalizability of the model to real-world [13, 17, 18]. Using DR has been proven to be very effective in transferring models in many tasks including robot manipulation, self-driving cars, and drone racing [13, 14, 15, 17, 18]. In the original works of DR, each randomization parameter that can control the appearance of the scene is uniformly sampled within a distribution range [13, 18]. One of the problems with randomizing all the parameters in the simulation environment from their distribution ranges is that we can end up with a lot of unrealistic environments [18, 22]. Unrealistic environments and wide randomization distributions can have a negative effect on policy training and cause infeasible solutions and hinder policy learning [22].

Accordingly, incorporating DR into synthetic data generation workflow to increase scene diversity while ensuring realistic configurations is a challenging task. Most of the data generation workflows follow a procedural model to insert objects into the scene which requires experts to manually tune parameters that govern the scene. To support DR through the manual procedural model to generate realistic and diverse scenes requires significant time and expertise to tune the parameters [14, 18, 19, 22]. Also, due to the error-prone and timely human effort involved, the generated data is limited in capturing the complexity and diversity of real-world situations [15]. Additionally, DR methods that are centered on self-driving car scenarios and situations [14, 15, 18, 19, 20, 21] are not generalizable to drone applications as the content gap in aerial imagery and dynamics of virtual camera motion representing drones is significantly different. ASDA generates realistic scene augmentations through a novel DR that is guided by scene graph information to automatically distribute objects in valid spatial arrangements over different types of simulation base maps.

### 2.4 Domain Randomization for aerial synthetic data generation

To increase diversity in aerial synthetic datasets used for training drones in visual Simultaneous Localization And Mapping (SLAM) tasks, TartanAir uses DR by utilizing AirSim [6]. TartanAir builds a diverse set of 30 pre-designed simulation scenes that cover different weathers, lighting and contexts that are fed to AirSim's data collection engine for collecting datasets with different complexities [7]. TartanAir has successfully shown that increased diversity in the datasets contributes to better performance and model transfer to the real-world for aerial navigation tasks. Additionally, in a recent study [29], it has been shown that the diversity of the TartanAir dataset can contribute to pretraining other navigation models that can transfer to new situations not seen in the training set which highlights the importance of incorporating DR in data generation workflow. But TartanAir's method of relying on carefully designed simulation environments is costly, limited, requires time and expertise, and is not scalable [7]. AirLearning [16] incorporates DR to the data generation workflow for reinforcement learning tasks by building on AirSim. AirLearning implements DR through configuration knobs that are randomly sampled from their distribution ranges and control assets' appearance and indoor environment features. The supported parameters are limited to object appearance and location for generating random obstacles in an indoor environment for reinforcement learning research. Accordingly, the method is not appropriate for large scale scene augmentation and doesn't generalize to other training tasks including aerial object detection. ASDA overcomes these limitations by procedurally generating unbound number of diverse realistic simulated environments with an extensive parameter support for aerial synthetic data collection and augmentation. Additionally, ASDA enables flexible control through a unified prompt-based interface which is important as many automatic methods provide little to no control [30].

## 3 ASDA: Aerial Synthetic Data Augmentation

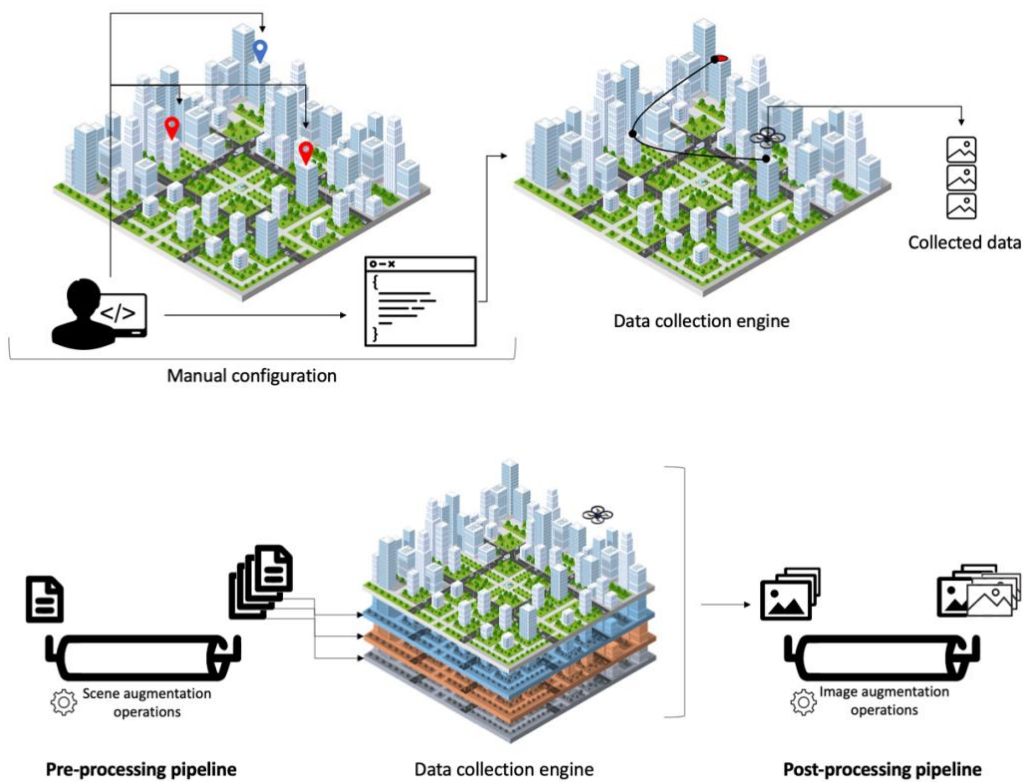

**Figure 1.** Procedural aerial synthetic data generation (Top) and (Bottom) ASDA procedural generative framework

ASDA is an aerial synthetic data augmentation framework to procedurally generate diverse realistic simulation scenes for aerial synthetic data collection. The framework extends a central data collection engine by adding two pre- and post- processing scriptable pipelines that automatically generate diverse simulation scenes and augment the generated data. ASDA replaces the manual configuration process involved in procedural models for aerial synthetic data generation with a generative approach for large-scale scene and image augmentation to increase end data diversity (Figure 1). ASDA provides a unified prompt-based interface over integrated pipelines that enables users to control pipeline operations through built-in action prompts and specify data generation strategy which increases workflow efficiency. To implement ASDA, we chose AirSim [6] as our central data collection engine.

### 3.1 Extending AirSim's Data Collection Engine

When collecting data through AirSim, there are three main steps involved: configuration, pose generation, and data collection. In the configuration step following a procedural model, parameters that govern the drone's trajectory and assets in an environment are manually tuned by an expert with specific values for the desired dataset that is to be collected. In the pose generation step, depending on the type of trajectory that is enabled and geolocations that are manually sampled from the simulation environment, AirSim's data collection engine would generate poses for the drone's trajectory across all geolocations that are specified. The simulation environment is rendered according to configuration and generated poses are sent to a virtual camera drone in the environment to record all required synthetic data (Figure 1). Data generated through this model lacks diversity and dependency on manual configuration creates challenges for large-scale diverse data generation. Additionally, data augmentation can only be performed after data is generated thus limiting diversity that can be applied to the data through large-scale scene augmentations before the data is collected through the virtual camera drone.

The pre-processing pipeline implemented in ASDA replaces the manual procedure of configuring each simulation environment supported by default in AirSim with a procedural generative method that automatically generates an unbound number of scene variations according to its operation specifications with an extensive parameter support to increase diversity in asset, environment, and drone features before feeding augmented scenes to the data collection engine. To further augmented the collected training data, ASDA also adds a post-processing pipeline to AirSim that has built-in augmentation operations that are applied over collected data. By connecting the two pre- and post- processing pipelines to the central data collection engine in AirSim, ASDA provides a unified data generation workflow with integrated augmentation operations to increase model generalization. Over the integrated data augmentation and data collection pipelines, we implemented a prompt-based interface for ASDA that provides a unified space for specifying data generation strategy by controlling built-in action prompts.

### 3.2 ASDA Architecture and Interface Design

To avoid generating unrealistic simulation environments through blind randomizations to increase scene diversity, we implemented a procedural generative approach for scene augmentation. Due to our procedure for scene augmentation based on a layered DR described in detail in following sections, we implemented ASDA with a pipeline-based approach, where implemented scene augmentation operations are added sequentially in order to generate a pre-processing pipeline. The pipeline uses scene specifications to seed its generative scene augmentation process where each scene specification is passed through the pipeline and each operation is applied to the specification as it passes through (Figure 2). Every operation has at minimum a probability parameter, which controls how likely the operation will be applied to generate each scene variation as its specification passes through the pipeline. Therefore, ASDA's pre-processing pipeline chains together operations that can be applied randomly. The parameters of each of these operations are also chosen at random, within a supported default range implemented for each operation and can be controlled by the user through the prompt-based interface. This means that each time a scene specification is passed through the pipeline, a different scene variation is returned. Depending on the number of operations in the pipeline, and the range of values supported by each operation, tens of thousands of new augmented scenes can be generated in this way. The post-processing pipeline follows a similar pattern where data augmentation operations are applied to each collected synthetic image as it passes through the post-processing pipeline after data collection to further augmented collected synthetic datasets (Figure 2).

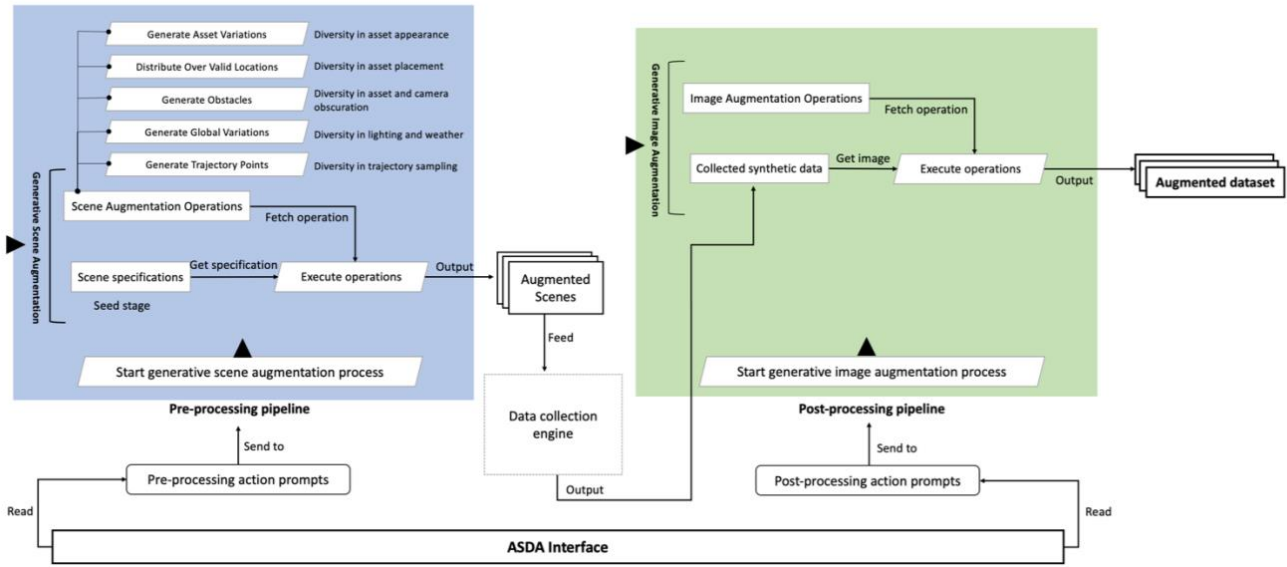

**Figure 2.** ASDA architecture and workflow

The operations in the pre-processing pipeline are implemented with a prompt-based interface architecture similar to the post-processing pipeline. Our reasoning behind having a prompt-based design was to provide a fluid control over generative approach of creating scene and data augmentations (Figure 3). This design enabled us to provide a unified interface for prompt-based data collection over the integrated pipelines so users can control the whole process in one place. The unified interface enables users to create end-to-end data generation strategies by specifying action prompts and without the need to manually interact with the simulator or configuration files. The pipelines then map specified action prompts to underlying pipeline operations. Table X shows a set of ASDA's action prompts with their usage.

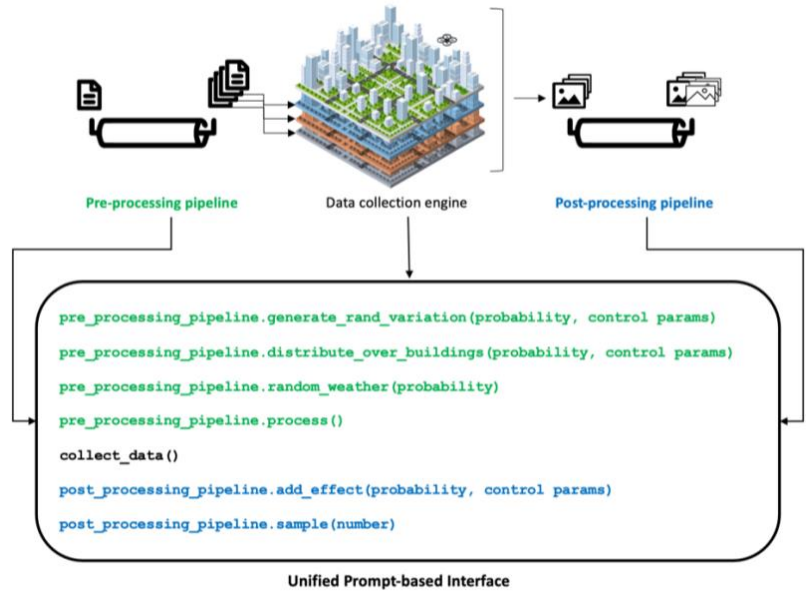

**Figure 3.** ASDA's unified prompt-based interface to control connected pipelines to data collection engine

Both pipelines in ASDA can be extended for adding new functionality. To extend the pipelines users can create a new operation by inheriting the base operation class and overwriting the perform method that has the operation execution logic with a custom code. Both pipelines have a method that receives a new operation and adds it to the existing pipeline operations.

### 3.3 Pre-Processing Pipeline

The pre-processing pipeline has built-in randomization operations for scene augmentation that can be combined and applied automatically to procedurally generate new variations of the simulation environment, drone agent, and trajectory parameters for data collection. The pipeline has 17 built-in randomization operations that can be added to the pipeline through the prompt-based interface. All operations can be customized by tuning control parameters to control automated randomizations applied over the simulation scenes. To automatically generate augmented simulation scenes by executing specified operations, the pre-processing pipeline has a generative scene augmentation process. The pipeline uses scene specifications to seed the generative scene augmentation process. Generative scene augmentation is enabled through a layered domain randomization (DR) approach incorporated into the pipeline in which the operations added to the pipeline are applied in separate layers on top of each other in order of precedence. Accordingly, randomization operations are categorized into five layers: asset variation generation, asset distribution, obstacle generation, global variation generation, and drone trajectory. Each category of operations is described in detail in the following subsections. By randomizing parameters within realistic distribution ranges for asset appearance, placement, obscuration, lighting, environment weather, and drone trajectory, ASDA automatically generates tens of thousands of variations from an initial scene specification for diverse data generation.

Given a scene specification which indicates the type of an existing simulation base map to use, ASDA's pre-processing pipeline automatically executes randomization operations in order of precedence to generate unbound number of augmentations of the input scene. If more than one scene specifications are inputted to the pipeline, the pipeline runs all operations over all specified scenes to generate augmented variations from each input scene. Users can classify the scenes and script the pipeline to only run certain operations over a certain class of scene specifications.

| Scenario | ASDA action prompts | Description |
| --- | --- | --- |
| 1) Generate 10 variations of the cellular tower asset for Seattle and San Diego scenes | pre_processing_pipeline.generate_rand_varation (1, scenes=["SEA","SD"], asset="cellular_tower", nvariations=10) | Here base configurations can be separated based on scene type: Seattle, San Diego. The pipeline reads the base specifications and labels them under scene types. The user can script the pipeline to call randomization operations that should be applied to all configurations without considering the class such as (1) & (2) and then can attach operations to be applied per class (3) & (4). |
| 2) Spawn them within 200-meter radius of center location of the cities | pre_processing_pipeline.distribute_asset_within_radius (1, mode="center", radius="200"] | |
| 3) Randomize weather over Seattle with higher probability for rain | pre_processing_pipeline.random_weather (1, scene="SEA", p=[0.6, 0.3, 0.1]) | |
| 4) Randomize weather over San Diego with uniform probability between rainy and sunny | pre_processing_pipeline.random_weather (1, scene="SD", p=[0.5, 0.5, 0]) | |

**Table 1:** Sample aerial synthetic data generation strategy using ASDA scene augmentation action prompts

### 3.3.1 Asset variation generation operations

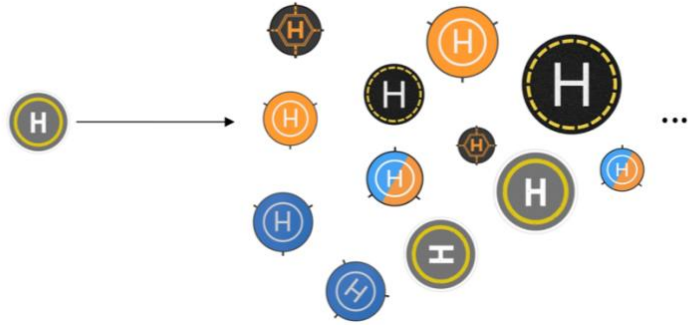

**Figure 4.** New variations of the landing pad asset to demonstrate function of asset variation generation operation

These randomization operations in the pre-processing pipeline generate new variations of the specified asset objects by randomizing their rotation, scale, and material to increase diversity in asset appearance (Figure 4). Both rotation and scale are randomized within a default uniform distribution range per X, Y, and Z axis. These operations can be customized by tunning control parameters passed to the operation function to only randomize a certain set of features of the object while keeping other features out of randomization. For example, the operation can be scripted to randomize scale while keeping the rotation the same. It can further apply rotation randomization to a certain axis specified by the user to overwrite the default behavior. The material randomization randomizes the color and texture of the asset objects. The textures are randomly picked from an array of texture files associated with each asset.

Additionally, we implemented an advanced shadow-casting operation to randomly cast shadows over assets to extend the variations that are generated per asset. The ability to capture shadows over objects in generated training datasets is important as many detection models used in drone inspection applications fail in the real-world when encountering different lighting situations due to dark shadows over objects [31]. One example of such a scenario is detecting cracks over wind turbines where the model for crack detection fails when there is a shadow affecting visibility of cracks over the turbines [32, 33].

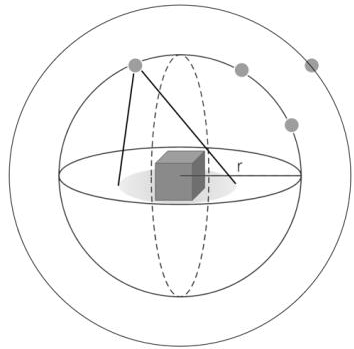

**Figure 5**. Demonstration of shadow casting operation for asset variation generation

To randomly cast shadows over objects in a simulation environment, the shadow casting operation automatically spawns a spotlight in randomly sampled location points from a sphere centered around the asset object (Figure 5). When sampling location points for the spotlight, the radius of the sphere is also randomized to further extend the distribution range for shadow's saturation as the light source is spawned closer or further from the asset object. The operation ensures that the generated randomized location for the spotlight is within the positive Y axis and re-generates the location if not. The result of randomized spotlight locations sampled over a sphere around a turbine asset is shown in Figure 6. The results are shown for both day and night global lighting conditions.

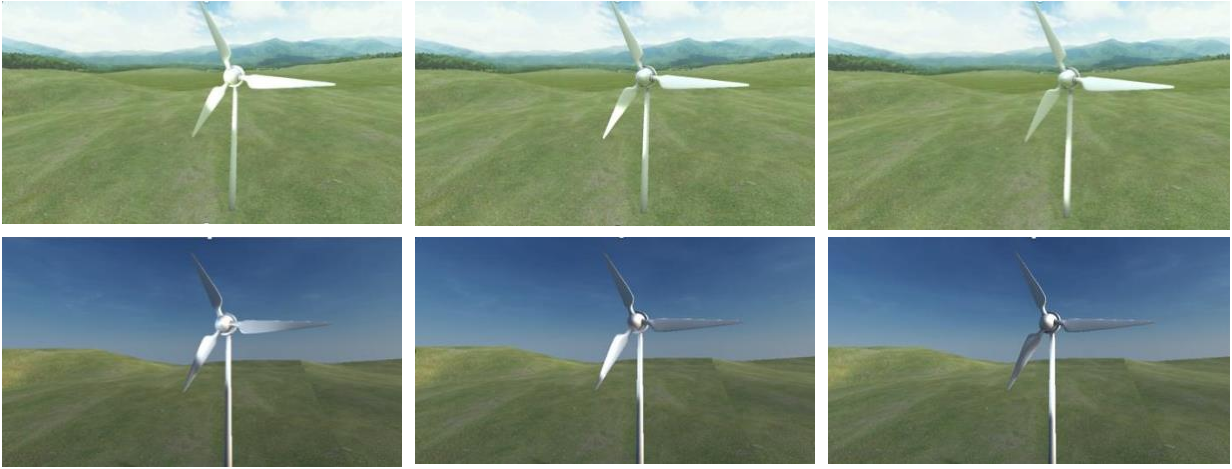

**Figure 6.** Result of running shadow casting operation in pre-processing pipeline to automatically generate different lighting conditions over the wind turbines

### 3.3.2 Asset distribution operations

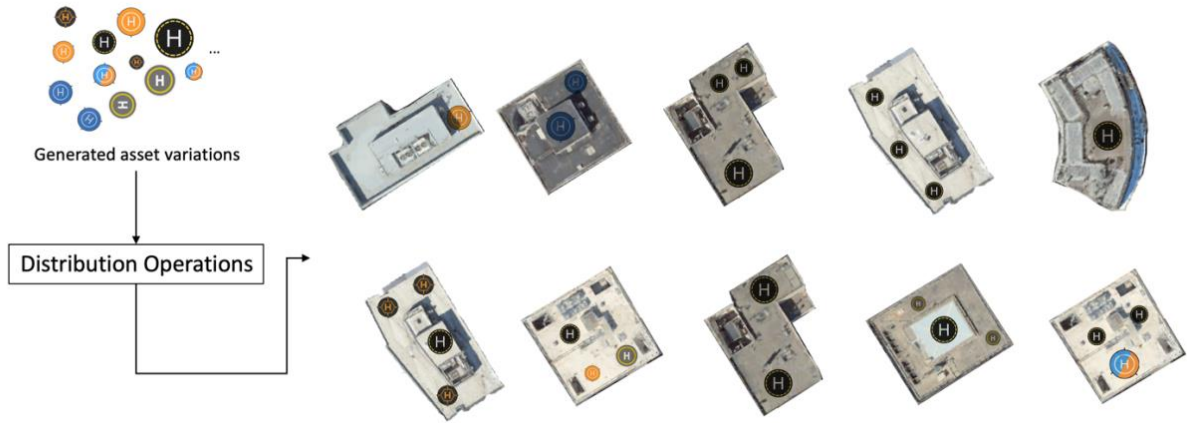

**Figure 7.** Sample result of applied asset variation and distribution operations to increase diversity of asset appearance and placement for scene augmentation

The goal of distribution operations in the pre-processing pipeline is to increase diversity of asset placements for scene augmentation (Figure 7). As mentioned in previous sections, the main challenge after generating asset variations for avoiding generation of unrealistic scenes is to find valid locations within the environment to randomly distribute the assets without dependency on real samples or manual workload. One advantage of drone-based scenarios for scene generation is the bird-eye viewpoint of the drone's camera over the environment map. Within the field of view of drones due to the larger area coverage, there are always parts of the environment's map that remain static, consistent and out of focus, especially for object detection tasks. Also, for navigation tasks, due to the altitude of the drone's trajectory over the map, randomizing scene objects for replicating challenging detect and avoid scenarios is easier as many structural features of the map remain static and ineffective on the training task. Accordingly, we can consider consistent parts of the environment in consequent frames in a drone's flight as a base map. For example, the terrain of wind farms is considered a base map for turbine distribution to train the drone for inspection task scenarios. With this perspective, we can reduce the cost of scene reconstruction object by object and instead shift the focus to distribution of objects that matter for the training task over these base maps. In the mentioned example, we can focus on generating and distributing the turbine objects over different base terrain maps. This is compared to the complexities of scene reconstruction for self-driving car scenarios where each object involved in the scene captured in the camera's field of view can affect navigation training based on its placement. Using this advantage and to create an adaptive automated method for asset distribution over any type of simulation environment map to generate new variations of it, we decided to embed 3D scene graph information of the environments to the distribution randomization operations in the pre-processing pipeline. A scene graph [34, 35] is a hierarchical model of 3D environments in form of a tree data structure that represents the spatial and logical relationship of a graphical scene. The tree contains a collection of graphics nodes including a root node representing the world in 3D simulation and a series of child group nodes that each can contain any number of child nodes and finally the leaf nodes serving as the bottom of the tree with no child. Each node in the scene graph contains transformation matrices that define the node's position in 3D space as well as other attributes of the node including its associated semantic label, bounding box, and material information.

The main benefit of scene graphs other than access to the hierarchical representation of environment objects is their traversability which enables querying the graph to find and extract particular objects of the scene. All the major game engines (such as Unity [36] and Unreal Engine [37]) use 3D scene graphs for scene representation which can be exported. We assume that the scene graph in the simulation is segmented so that the group nodes representing distinct objects in the scene have a semantic class (tree, building, road, …) in addition to position and bounding box attributes that are supported by default. Accordingly for distributing each asset type, the operation queries the extracted scene graph to find places over which the asset can be spawned over at the time of rendering. Querying the scene graph per asset type results in extracting a sub-graph that we call *distribution space* which includes all the nodes representing the container objects (e.g. places) over which the asset can be distributed over. To further reduce the subgraph to the boundary boxes and attributes that directly inform location sampling for spawning assets, we map the sub-graph to a distribution layer that includes the boundary boxes within the container objects together with relevant attributes for spawning. Accordingly, randomized variations of each type of asset are spawned over valid locations sampled from free spaces mapped from the sub-graph on the distribution layer. This is demonstrated in Figure 8 with an example of distributing a landing pad asset over an environment. The distribution space sub-graph first includes all the building objects representing places that the landing pad can be spawned over. Then the distribution space is mapped to a final distribution layer that only includes boundary boxes that are polygons of the buildings' rooftops as well as the relevant attributes such as the height of the buildings and their type. Accordingly, the distribution layer informs the operation to sample locations within the boundary boxes representing free spaces to spawn the asset. The additional attributes such as the building's type in this case enable distribution operations to support more complex scenarios if needed by performing filtering over distribution layers based on attributes associated with each boundary box. These operations also support a pattern specification that allows random location sampling within spaces in a specific pattern over the distribution layer. The patterns include location sampling within a radius, line or a polygon within the boundary boxes.

Due to our goal for scalability and adaption of scene augmentation to different types of simulation environment maps, we took a step further to make sure our method can be applied to different types of environments. Currently, there are three types of environments used in simulators for synthetic data collection: designer 3D environments, NeRF blocks [24], and 3D maps. Both designer 3D environments and NeRF blocks provide access to the scene graph representing the scene, but the scene graph cannot be extracted from 3D maps. 3D maps are imported from third-party applications that provide 3D mapping services and geo-data based on satellite imagery thus access to a graphical representation of objects in the scene is not supported when these maps are imported in the simulation environment. With increasing applications of drones for UAM[1], drone delivery, and cellular tower inspections, the need for using 3D maps of cities to increase the speed for generating synthetic datasets are on the rise. Additionally, these maps provide a high-fidelity representation that is hard and time-consuming to implement using other approaches such as NeRF or manual design. Accordingly, it's important to extend the usability of our asset distribution as a critical layer of ASDA's DR and scene augmentation process to support 3D maps. Additionally, by supporting 3D maps, our scene augmentation operations can be used in situations where ground-truth annotated scene graphs are not available. Accordingly, we implemented a workaround for embedding scene graph information into distribution operations for 3D maps. Instead of relying on scene graph information when using these maps, we fetch the available labeled structural information of these maps by querying OpenStreetMap [38] data by embedding OverPass [39] to randomization operations. OverPass is a geographic data mining tool for OpenStreetMap data that has a query language for querying and filtering the map data using location, type of objects, tag properties, proximity, or combinations of them which is close to querying an annotated scene graph in a 3D environment. Extracted data includes similar information including boundary boxes of the roads, buildings, and other manmade structures in the maps together with their coordination and attributes such as height, type, name, address, etc. By incorporating OverPass into ASDA's randomization operations, we implemented a backbone for asset distribution over 3D maps to rewrite the default behavior of reading the information from the scene graph. As asset distribution is a critical layer of ASDA's DR, supporting different environment types in this layer makes its scene augmentation environment type-agnostic meaning that it can generate randomized variations of any type of base map, whether a carefully curated 3D scene, a 3D map or a NeRF block, without needing to change data generation procedure.

---

[1] Unmanned Air Mobility

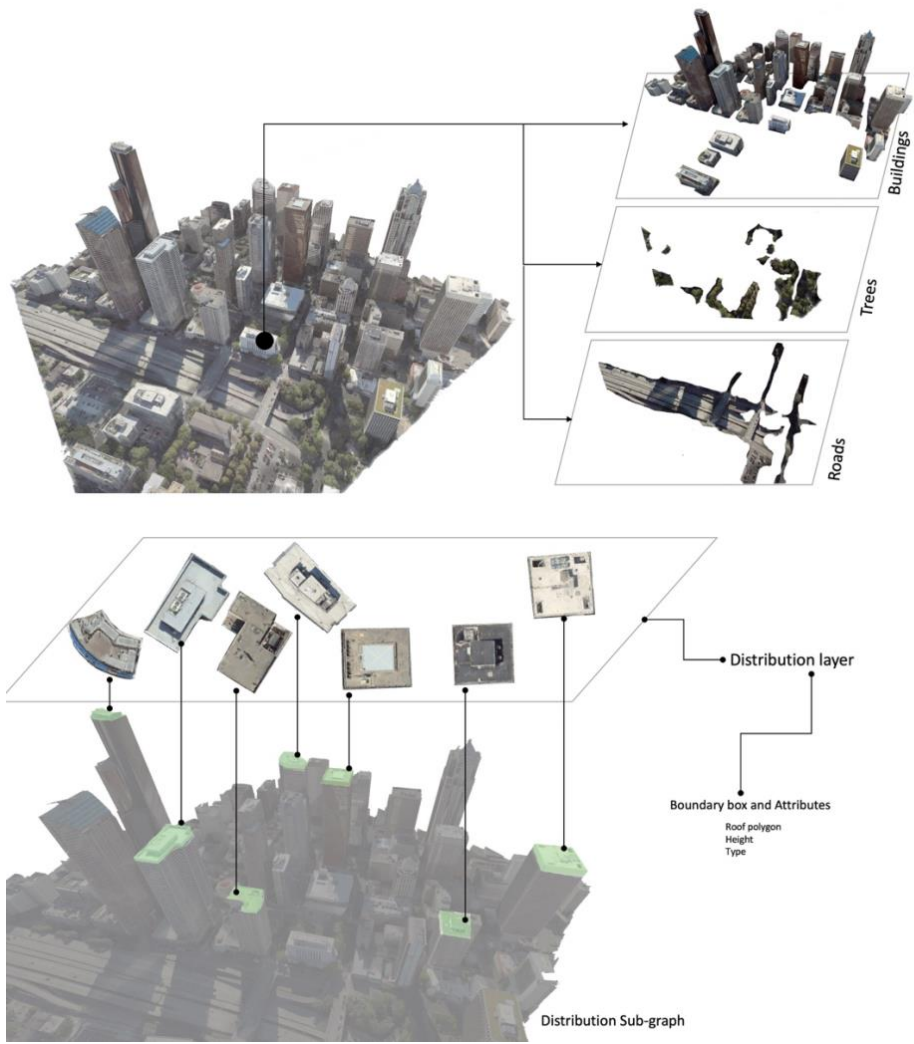

**Figure 8**. Demonstration of distribution layer mapping from the scene graph for a landing pad asset distribution over the environment. The sub-graph includes nodes representing building objects and their attributes that are separated from the main graph and then mapped to the distribution layer including the buildings' roof polygons as boundary boxes with relevant attributes that represent free spaces to inform location sampling behavior for spawning the asset.

### 3.3.3 Obstacle generation operations

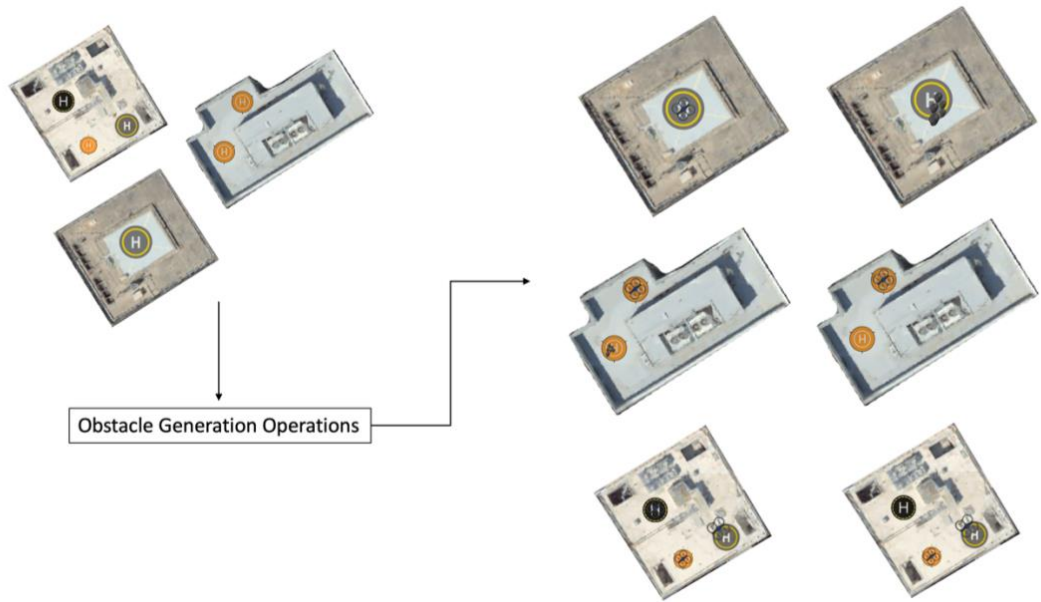

**Figure 9.** Sample result of applied obstacle generation operations to increase diversity of asset and camera obscuration

The goal of obstacle generation operations in the pre-processing pipeline is to increase diversity of asset and camera obscuration for scene augmentation (Figure 9). As one of the main challenges in vision-based navigation and detection for drones is camera and object obscuration, we implemented a set of operations in pre-processing pipeline to replicate such situations in scene augmentations. Accordingly, we implemented two types of randomization operations for obstacle generation: asset obstacles, and Field Of View (FOV) obstacles. Asset obstacle generation operations generate random variations of an obstacle object over a desired asset by sampling their locations over a sphere centered around the asset (Figure 10). The obstacle generation over assets can work as noise generation to make the main asset less visible in different random patterns.

Additionally, to replicate drone camera obscuration situations, we implemented an advanced operation to generate random obstacles within the FOV of the drone's camera. Accordingly, we can recreate situations where another drone is in the FOV of the current drone for detect and avoid training, we can create different situations of limited FOV by randomizing obstacles in different distances of the camera, and many other scenarios useful to the vision-based navigation and detection training tasks.

The implemented FOV obstacle operation samples points from the conic area that represents the FOV of the drone's camera based on its pose within a random distance of the drone. It will then spawn random variations of the obstacle objects in the sampled points based on their distance from the camera (Figure 10).

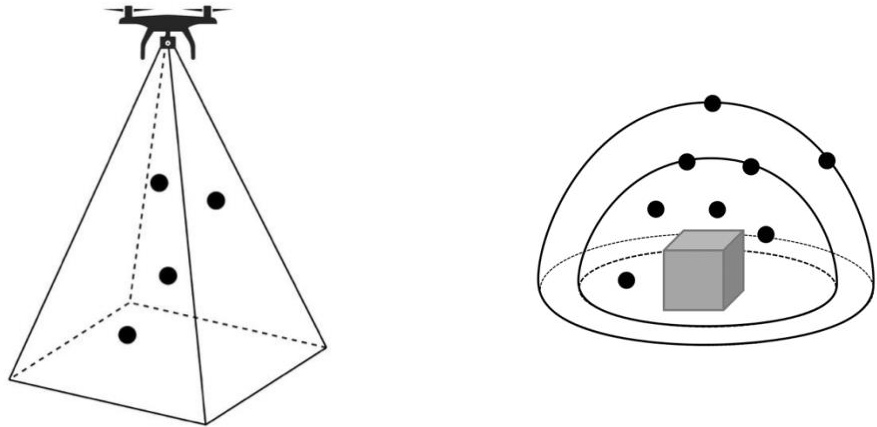

**Figure 10.** Two types of obstacle generation. Left: FOV obstacles; Right: asset obstacles

### 3.3.4 Global variation generation operations

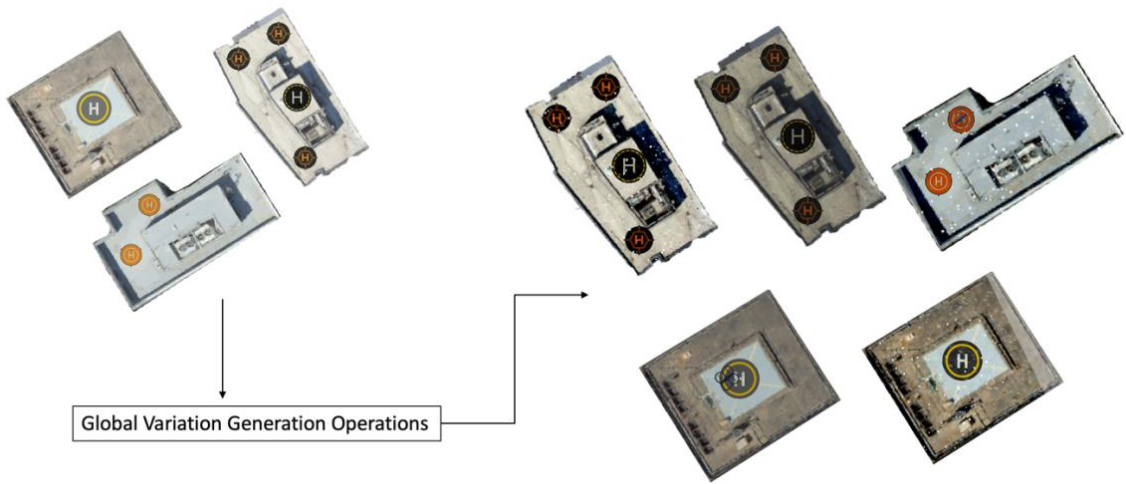

**Figure 11.** Sample result of applied global variation generation operations to increase diversity of environment light and weather conditions

These operations randomize light and weather controlling the global appearance of the simulation scenes after assets and obstacles are spawned over the scene (Figure 11). The global parameters include weather, time and lighting. Each global parameter is uniformly randomized within a default distribution range, but the operations are implemented such that they can be scripted to accept probability per global parameter value to rewrite the default behavior. With probability distribution, there is more flexibility in global situations captured in the generated datasets based on the type of scenes or training task needs. For example, in the scenario described in Table 1, we can increase the probability of rainy weather over the Seattle scene to inject more context-aware specifications into the generated datasets.

### 3.3.5 Drone trajectory generation operations

These operations are applied in the last layer of ASDA's DR in the pre-processing pipeline for the drone's trajectory pose generation before feeding augmented scenes to the data collection engine. There are two categories of these operations implemented, one targets the randomization over trajectory's location points and the other over the trajectory specifications. Operations that randomize location points for trajectory, randomly sample desired number of locations from the environment and add it to the drone's trajectory. By default, all the locations of main assets distributed and spawned in the scene are added as a trajectory point for feeding to data collection engine. When the location sampling operations are invoked, the sampled locations are added as additional points for the trajectory. There are multiple types of location sampling supported including sampling locations from the embedded scene information, as well as sampling locations based on a pattern (within a radius, line or polygon). The operations targeted at randomizing trajectory specifications, randomize trajectory patterns as well as drone's camera pose and motion by randomizing pitch, yaw, and roll and the distance within the trajectory that defines the data capture rate.

### 3.4 Post-processing pipeline

The post-processing pipeline has 33 built-in augmentation operations and is applied after the data is collected through AirSim data collection engine over the augmented scenes generated by the pre-processing pipeline. The goal of the post-processing pipeline is to further augment the generated datasets by applying image augmentation operations over collected synthetic images (Figure 2). This pipeline is implemented over the Augmentor python library [40] for image augmentation. The pipeline reads the images generated by the data collection pipeline and applies augmentation operations such as random rotation, flipping, erasing, zooming, etc. to the images. There are 32 operations in the library that are incorporated, and we also implemented an additional operation for the pipeline that enables applying visual effects as overlay layers to the images to perform advanced operations that are otherwise impossible or highly render expensive to generate in 3D simulation. One such operation is simulating fog and clouds to replicate low visibility situations that affect navigation and detection. We implemented the visual effect augmentation operation to create similar situations by merging effect layers on top of generated images. The results are demonstrated in our experiment (Section 4).

### 3.5 Data Generation Scenarios based on ASDA

To better demonstrate the potential of ASDA, Table 2 includes few sample data generation scenarios that can be specified through the unified prompt-based interface of ASDA and executed by the pipeline by running the operations to generate desired datasets.

| **Data Generation Scenarios** | **ASDA Action Prompts** | **Use Case** |
| --- | --- | --- |
| - Generate 20 variations of a wind turbine asset with different crack textures and with different rotations in x and y axis, and different scale in y axis<br>- Distribute turbines within 2 km radius of center of all wind farm terrains<br>- Randomize shadow casting over turbines<br>- Randomize weather in the environment<br>- Augment images with random erasing | pre_processing_pipeline.generate_rand_varation (1, scene=["WINDFARM "], asset="wind_tourbine", nvariations=20, rotation_axis=["x","y"], scale_axis=["y"])<br><br>pre_processing_pipeline.distribute_asset_within_radius (1, mode="center", radius="2000"]<br><br>pre_processing_pipeline.random_shadow (1, asset="wind_tourbine")<br>pre_processing_pipeline.random_weather (1)<br><br>collect_data()<br>post_processing_pipeline.random_erasing (0.5) | * Crack detection over turbines<br>* Detect and avoid for navigation and path planning |
| - Distribute 10 cellular towers over all land site scenes within 12 miles of each other<br>- Generate and randomize up to 4 obstacle drones around the cellular towers<br>- Generate Cylindrical and Point2Point Trajectory over all towers | pre_processing_pipeline.generate_rand_varation (1, scene_class=["city"], asset="cellular_tower", nvariations=10)<br><br>pre_processing_pipeline.distribute_asset_over_area (1, distance="12"]<br>pre_processing_pipeline.sample_location (1, types=["cylindrical","point2point"])<br>collect_data() | * Detect and avoid for navigation and path planning |

| Data Generation Scenarios | ASDA Action Prompts | Use Case |
|---|---|---|
| - Sample 10 trajectory points from buildings over San Diego and Seattle<br>- Generate obstacle objects in FOV of the drone to randomly obscure the camera<br>- Randomize weather over all environments<br>- Randomize camera pose, drone's altitude, and motion to generate trajectories over sampled geolocations<br>- Augment images with random erasing | pre_processing_pipeline. sample_location (1, scenes=["SEA","SD"], asset="building", npoints=10)<br><br>pre_processing_pipeline.random_obstacle_in_FOV (1, obstacle_asset="obstacle_drone")<br><br>pre_processing_pipeline.random_weather (1)<br><br>pre_processing_pipeline.random_trajectory (1, features=["camera_pose","altitude","motion"])<br><br>collect_data()<br>post_processing_pipeline.random_erasing (0.5) | * Optical flow generation<br>* Visual Odometry |
| - Sample trajectory points from all city maps<br>- Randomize weather, lighting, and time over all scenes | pre_processing_pipeline. sample_location (1, scene_class=["city"])<br>pre_processing_pipeline.random_weather (0.6)<br>pre_processing_pipeline.random_time (1)<br><br>collect_data() | * Localization |
| - Generate 2 variations of the landing pad<br>- Distribute them over 10 restaurants over all city maps<br>- Generate random drone obstacles over the landing pads<br>- Randomize weather, and lighting<br>- Augment the images with random rotate, and cloud visual effect | pre_processing_pipeline.generate_rand_varation (1, scene_class=["city"], asset="landing_pad", nvariations=2)<br>pre_processing_pipeline.distribute_asset_over_amenity (1, type="restaurant", number=10)<br><br>pre_processing_pipeline.random_obstacle_over_asset (1, asset="landing_pad", obstacle_asset="obstacle_drone")<br><br>pre_processing_pipeline.random_weather (0.6)<br>pre_processing_pipeline.random_time (1)<br><br>collect_data()<br><br>post_processing_pipeline.rotate(probability=0.7, max_left_rotation=10, max_right_rotation=10)<br>post_processing_pipeline.add_effect(probability=0.8, effect_name="visibility", intensity=1) | * Landing pad detection<br>* Detect and avoid for navigation |

**Table 2:** Example data collection scenario scripts using ASDA operations and their possible use case

## 4 EXPERIMENT AND RESULTS

To evaluate ASDA and validate its functionality, we ran an experiment for generating data for a landing pad detection task. For our experiment, we used the Seattle simulation scene. The Seattle environment is one of the 3D city maps provided by Bing maps [41] which is supported by AirSim (Figure 12) and we use it to seed the generative scene augmentation process in ASDA's pre-processing pipeline. We chose to test our method over a 3D map to evaluate the adaption of our domain randomization method over environments with no direct access to the scene graph for asset distribution. Figure 13 shows the experiment result and the difference in end data after extending AirSim's central data collection engine with ASDA pipelines and replacing AirSim's procedural model for configuration to generate large-scale diverse datasets.

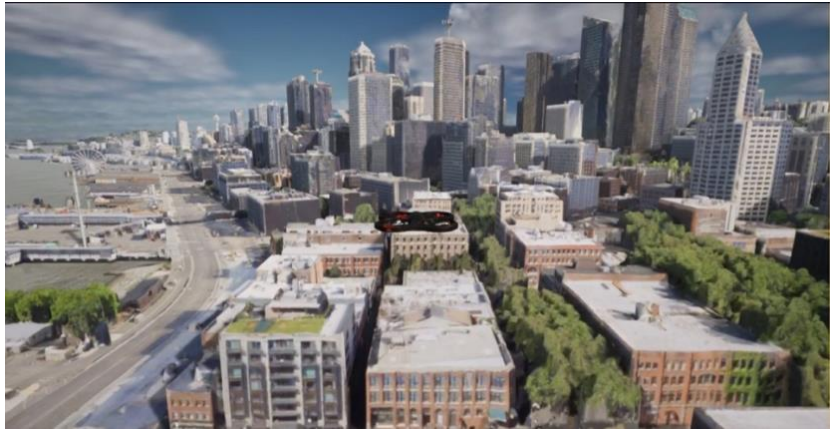

**Figure 12.** AirSim's simulation environment with the Seattle 3D map

ASDA's pipelines for the experiment were scripted through interface action prompts to have the following data generation scenario executed: 1) generate 5 variations of the landing pad (by default rotation, scale and material are randomized) 2) spawn the landing pads over 7 different buildings 3) add noise over landing pads by randomizing particles within 2-meter radius of spawned landing pads 4) randomize other drones in the FOV of current drone 5) randomize drones over spawned landing pads 6) randomize multiple landing pads over same buildings 7) randomize lighting 8) augment the generated synthetic images by random rotation, flipping, zoom and distortion 9) augment the images with visibility effect.

Before extending AirSim's data collection engine with ASDA, the procedure involves manually running a simulation instance to find and specify each building's geolocation by an expert user then tunning relevant environment parameters in a configuration file. Asset variations should be manually generated and specified. To distribute each asset over each building, the user needs to run 5 separate instances (for each of the 5 variations in this case) of the simulator with each configured to spawn each of the assets over specified locations. Other scene augmentations wouldn't be possible in the manual procedural model or extensively time-consuming. All the manual workload in the procedural process is automated through ASDA's pipelines and the unified interface thus increasing data generation efficiency by 50%. Additionally, through the extensive scene and data augmentation support in ASDA and ability to generate unbound number of scene variations before data collection, ASDA increases end data variants by more than 75% with its current 50 built-in operations enabling large-scale scene augmentations. In our experiment, more than 150 synthetic RGB images with realistic diverse configurations are generated automatically (Figure 13). It's important to note that to validate the operations and evaluate whether ASDA can produce a diverse dataset, we bounded the number of generated data in the pipelines to no more than 150 images, but the pipelines can be scripted to generate any number of desired scene and data augmentations for a training task.

The results confirm the potential of ASDA's data augmentation framework in automatically generating realistic and diverse datasets that capture a wide range of settings that can occur in real-world situations. It further shows the benefit of allowing control over the automated procedure of operations and the parameters they govern in the scripting process. Additionally, the results validate the adaptability of our domain randomization approach to publicly available 3D maps to increase the scale and speed of data augmentation for advancing the generalizability of trained models while minimizing dependencies on human effort and real samples in the procedure.

**Original dataset**

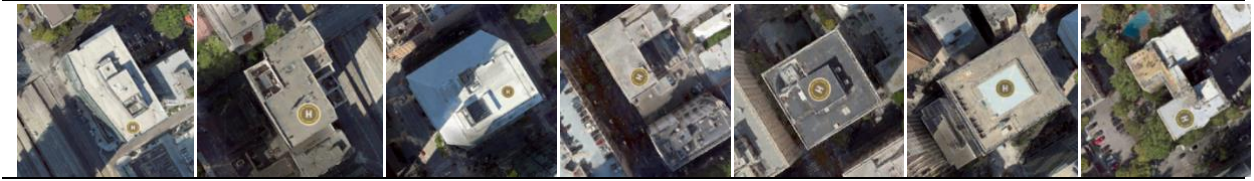

↓ ASDA

**Pre-processing Scene Augmentations**

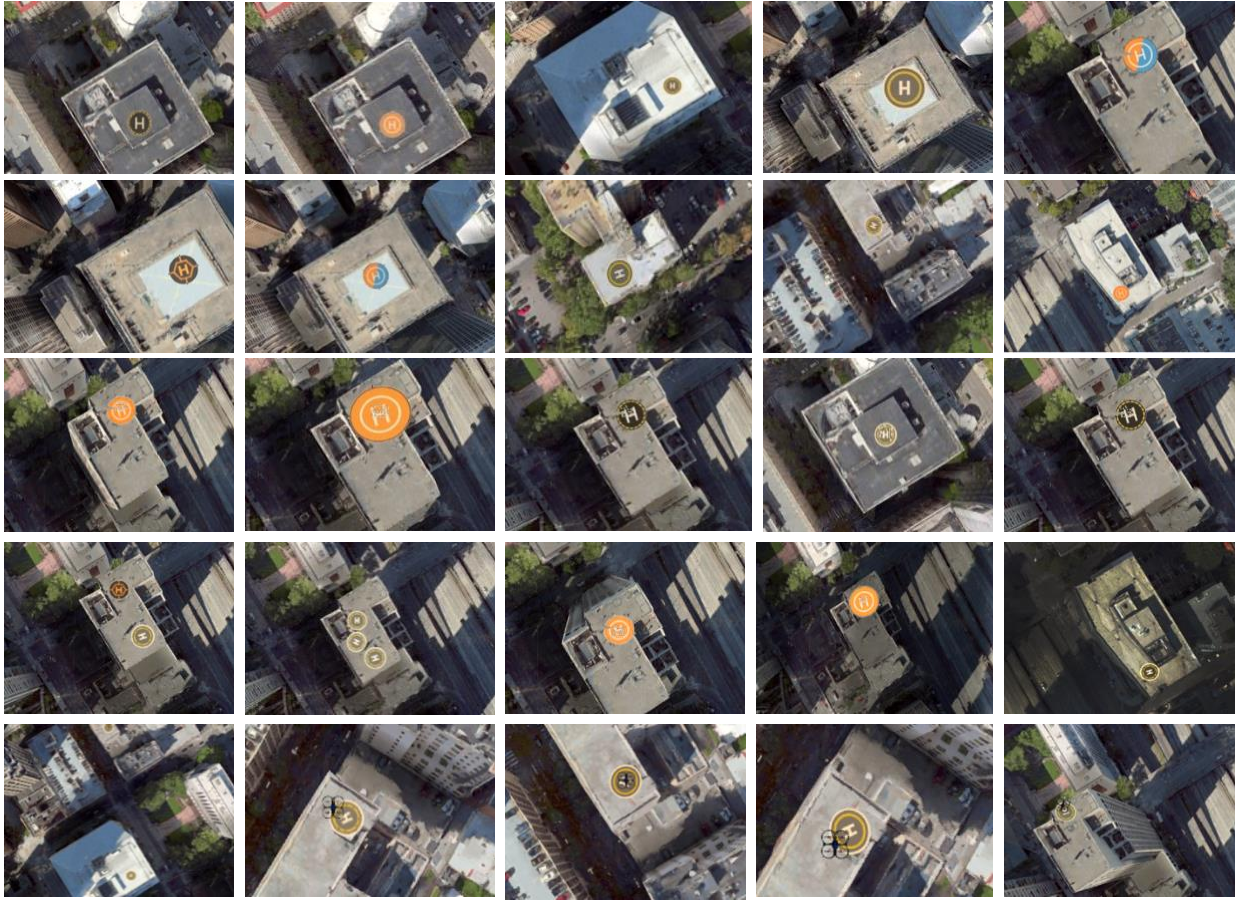

**Post-processing Image Augmentations**

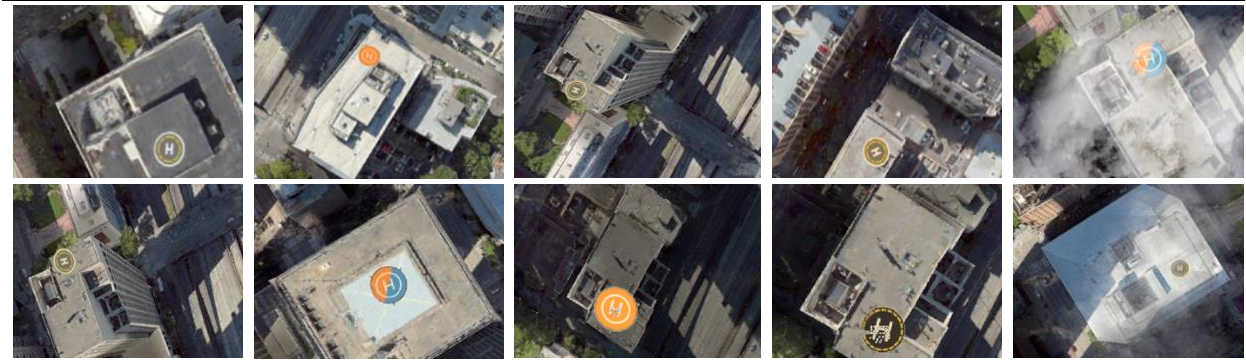

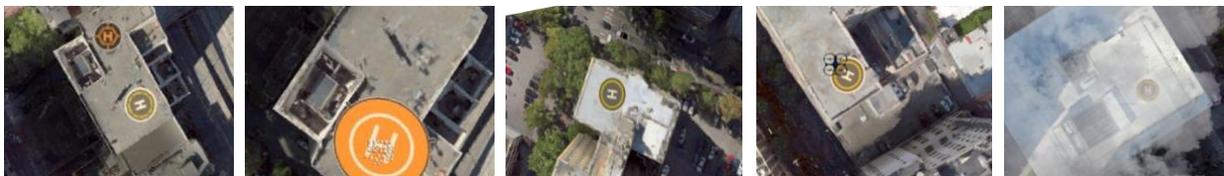

**Figure 13.** Sample from generated RGB synthetic dataset for landing pad (LP) detection. Scene augmentations resulting in diversity of LP material, placement, light conditions, noise generation, obstacle creation over LP. Image augmentations showing random flipping, distortion, zoom, rotation, and cloud effect augmentation.

## 5 DISCUSSION

Here we discuss the design benefits, usability and applications of ASDA's framework and opportunities that can be taken to further advance our method by highlighting the challenges and future directions of the current work. Additionally, we propose a data collection optimization based on ASDA to maximize the trained model's accuracy and performance on downstream task while minimizing costs.

### 5.1 Procedural Generative Approach to Synthetic Aerial Data Augmentation

As we argued at the beginning of the paper, the scalability of the data augmentation method is very important to support different scenarios and training task needs with high degrees of flexibility to adapt. While minimizing human effort was our purpose to have an automated way of generating diverse datasets, it's important to allow a certain level of control and procedure support [18] so that the users can customize the data augmentation by adding context-specific information, adapting specific strategies, and determining the degree of complexity and randomization that is needed. With ASDA's modular design of pre- and post- processing pipelines and the unified prompt-based interface, users can easily adjust and control the degree of automation and generative functions that are executed in the end-to-end data generation. Accordingly, they can combine different operations to implement desired strategies, they can control the customization over distribution ranges to overwrite default automated behaviors, and they can group operations and apply them over a batch of scene specifications at once to scale the strategies for data generation operations for training tasks with similar needs. Additionally with the layered design of the incorporated domain randomization (DR), not only users can benefit from the adaptive DR to different types of base maps as mentioned in the paper but also it gives users the benefit of deciding on the layers to combine and have depending on the training task needs. For example, for object detection tasks, obstacle generation and object variation layers might be centered while for a localization task, the trajectory layer is more important, and users can remove object variation from the DR layers. The design also allows adding additional layers to the DR by adding customized operations to the pipelines. With the embedded scene graph information in the randomization operations, users can implement more complex and advanced strategies for data collection over the scenes. Additionally, ASDA supports generating diverse datasets for both navigation and detection tasks while previous methods mostly focused on either of these tasks. Accordingly, we can generate datasets for detect and avoid, object detection, localization, and visual odometry all with the DR adapted based on the task need over environments. By implementing automated operations and scriptable pipelines that can be used in different ways without the need to manually tune parameters by running simulation instances or relying on real samples, we significantly cut the cost of data collection while increasing the scalability and flexibility of the augmentation method for a diverse set of training tasks.

### 5.2 Using Scene Graph to Guide Domain Randomization

We already covered the benefits of embedding the scene graph information into randomization operations to distribute objects in valid configurations over different types of environments in section 3.1.2. But one remaining challenge in our work is that we incorporated explicit human knowledge in our implemented operations to map the scene graph to a distribution layer for each asset type. Accordingly, the type of container objects (places) that can be used for spawning asset objects and should be extracted as a sub-graph are determined by the user when scripting the operation. For example, for the landing pad object, the user calls the distribution operation with *building* type thus guiding the mapping operation to create the sub-graph that contains the nodes representing building objects. But to further scale this process, we can train a model to learn the spatial relationships between different types of objects from publicly available images to then use the learned relationships to guide the operation for associating each asset type to the type of container objects (places) from the environment that it can be distributed over. This would help in further having a single distribution operation that can take the asset with its label and determine the types of objects that should be extracted in the distribution space sub-graph with relevant attributes. In the landing pad example, the operation would take the landing pad asset and automatically generate the distribution space sub-graph with building objects and their attributes from the scene graph without receiving an input for the semantic category of the distribution nodes which is "building" in this case. To train a model so that it can learn the spatial relationship between objects in an image, we can adapt methods such as the one discussed in [34] that show how we can extract the spatial relationship between types of objects from scene graphs generated over 2D images.

### 5.3 Data Collection Optimization based on ASDA

ASDA's pipelines support augmentation operations that can generate many variations of simulation scenes for collecting infinite amount of diverse aerial synthetic datasets, two important questions remain: *how much data should be collected* and *how many variations per environment scene should be generated* for each type of training task. Optimizing data collection for sim-to-real transfer proposes one major problem and that is to determine whether we need to increase data size with the same scene setting and complexity to meet performance targets for a model or whether we need to increase the complexity and diversity of the scenes captured in the data by generating more variations of the scenes. To propose a solution, we first reflect on two important works on data collection optimization from which we get inspiration from. Mahmood et al. [42] suggest a Learn, Optimize, Collect (LOC) optimization strategy for data collection in which the cost of data collection to meet a performance target ($V^*$) is minimized through an iterative data collection approach. Accordingly, the data is collected in multiple iterations and feedback is received on the model's performance per iteration to re-evaluate and decide how much more data to collect while minimizing risk of failing to meet $V^*$ (the performance target) in the next iteration until the target is reached. To optimize the iterations, they estimate the probability distribution of how much data is needed by collecting performance statistics over sampled data and estimating data requirements. In LOC, authors model performance as a function of data set size thus the complexity of the settings captured in the data is not considered in the strategy. Mikami et al. [43] suggest that increasing dataset size is ineffective when the performance target is not being reached due to the content gap in sim-to-real scenarios where the pretraining is done with synthetic images and the model is fine-tuned on real images. Accordingly, they propose a scaling law (1) for explaining generalization error on fine-tuning in sim-to-real transfer and suggest estimating the parameters of the law (e.g. $C, \alpha$ where $C$ is the transfer gap and $\alpha$ is pretraining rate) helps to judge whether we should increase the data size or change the setting of image synthesis based on performance feedback.

$$Test\ error \cong Dn^{-\alpha} + C \qquad (1)$$

Inspired by both works, we suggest a combination of these methods for optimizing data collection based on ASDA's architecture. Our solution is built on the LOC strategy but modified based on ASDA and adjusted to address the limitations for sim-to-real according to Mikami et al.'s scaling law. Accordingly, we extend the estimation model used in LOC to estimate the parameters of the scaling law (1) based on the performance feedback received at each iteration. Accordingly, the model performance is a function of data size as well as the content gap to determine whether we need to change the setting and complexity or merely increase the data size for the next iteration.

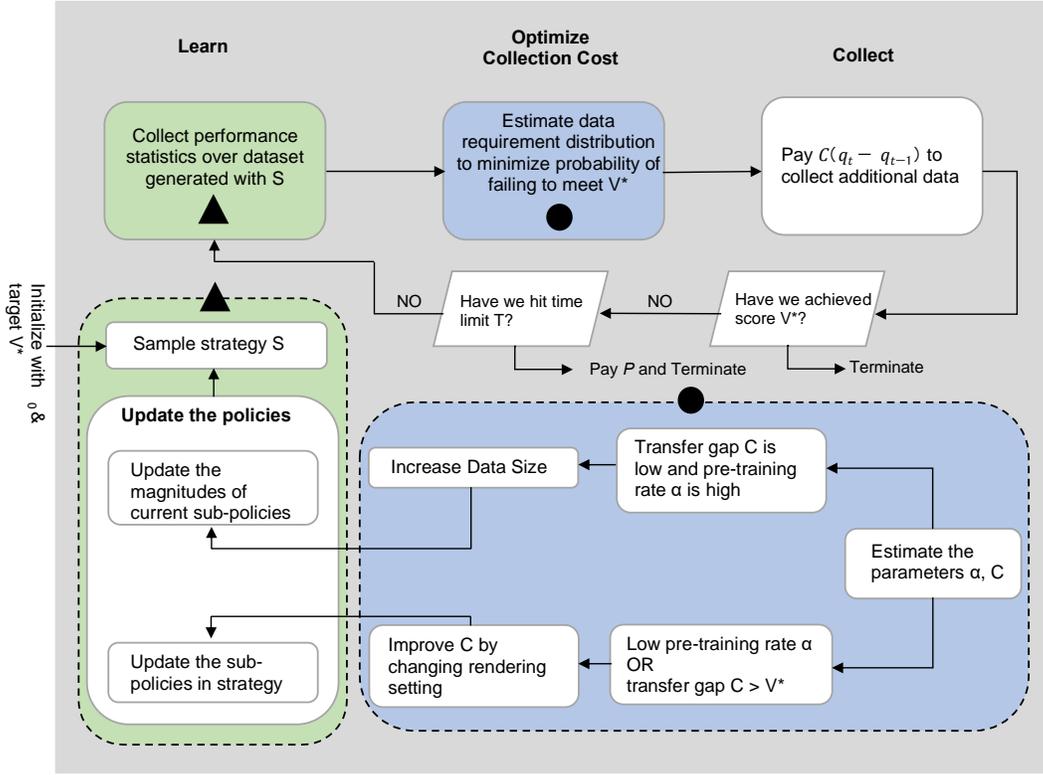

**Figure 14.** Proposed data collection optimization solution based on ASDA

At each iteration depending on the decision, we need to apply appropriate changes to the data generation pipelines to increase the size or diversity of the data before the next iteration. The complexity and diversity captured in the scenes are affected by randomization and augmentation operations used in the pipelines as well as the distribution ranges per parameter that the operations work with. The size of the data is affected by the configuration of pre-processing scene augmentation operations (for example distributing over all buildings in a city map or sampling 10 buildings affects the size of data generated) as well as the sample size specified for image augmentation operations. Accordingly, to learn the best strategy to adapt to improve the generated data, both operations and the parameter distribution ranges need to be optimized based on feedback from the previous iteration to optimize the data generation strategy for next iteration.

Inspired by Cubuk et al.'s work [44] on choosing the best augmentation operations for image classification tasks, we formulate the problem of finding the best randomization and augmentation policies as a similar discrete search problem. The search algorithm is incorporated in the Learn component of LOC where it samples a data randomization and augmentation policy S, which has information about what operations to use, the probability of using each operation, and the magnitude of the operation (distribution ranges of parameters affected by the operation). Accordingly, policy S will be used to train a neural network whose validation accuracy is $V^*$ that is sent back to update the controller of the network. In the search space, a policy (data generation strategy S) consists of one or more sub-policies with each sub-policy consisting of operations that need to be applied in a sequence. For example, all generation operations for obstacle or asset variations are followed by a distribution operation.

Accordingly, at each iteration the dataset $D_{qt}$ ($q_t$ is data points collected at the end of iteration T in dataset $D$) is generated based on the strategy S, and S is optimized based on performance feedback received at each iteration $V(D_{qt})$ (where $V(D)$ is score function) until the performance target $V^*$ is reached. To optimize S, estimated parameters $C, \alpha$ determine the decision for increasing data size or changing the diversity and complexity of the scene. If the transfer gap $C$ is low and the pre-training rate $\alpha$ is high, we need to increase the data size. If the pre-training rate $\alpha$ is low or $C$ is greater than target performance ($V^*$) then we need to improve data by changing the rendering setting. Based on the outcome, the policy in S is updated to optimize for next round. Accordingly, if data size change is needed, only the magnitudes of the current sub-policies meaning the distribution ranges of parameters affected by operations in current sub-policies are updated for the next iteration but if change in rendering setting is needed, the sub-policies themselves are re-sampled and changed in the strategy S (Figure 14).

Using the proposed data collection optimization strategy over our ASDA's data generation scheme can reduce cost and development delays for generating the datasets and results in generating more advanced and high-quality benchmark datasets for different training tasks. Additionally, the strategies that are learned per training task can be outputted separately to inform users of the best policies to deploy for their data collection needs. For example, a set of suitable operations for maximized performance in landing pad detection with refined distribution ranges can be offered for pretraining detection models.

### 5.4 Integrating LLMs with ASDA for Data Generation

ASDA's prompt-based interface design over data augmentation pipelines can contribute to advanced aerial synthetic data generation workflows. ASDA's prompt-based interface for the pipelines can be integrated with Large Language Models (LLM) such as OpenAI's ChatGPT [45] to support natural language for more fluid data generation experience. ChatGPT can combine text generation with code synthesis enabling new ways of interaction through dialogue for code generation. Accordingly, by linking the library of ASDA's pipeline action prompts to ChatGPT, users can indicate their data generation goal for a training task and ChatGPT can parse user intent and convert it to pipeline action prompts to write the data generation strategy (Figure 15). By using our proposed pipeline for data collection optimization (section 5.3) in the data generation loop as feedback (Figure 15), the data generation strategy generated by the LLM can be automatically improved by revising policies to apply targeted augmentations for improved generalization. This strategy would further reduce the user intervention in data generation process and can further increase the efficiency of the workflow. We intent to implement this strategy in our future work to improve ASDA.

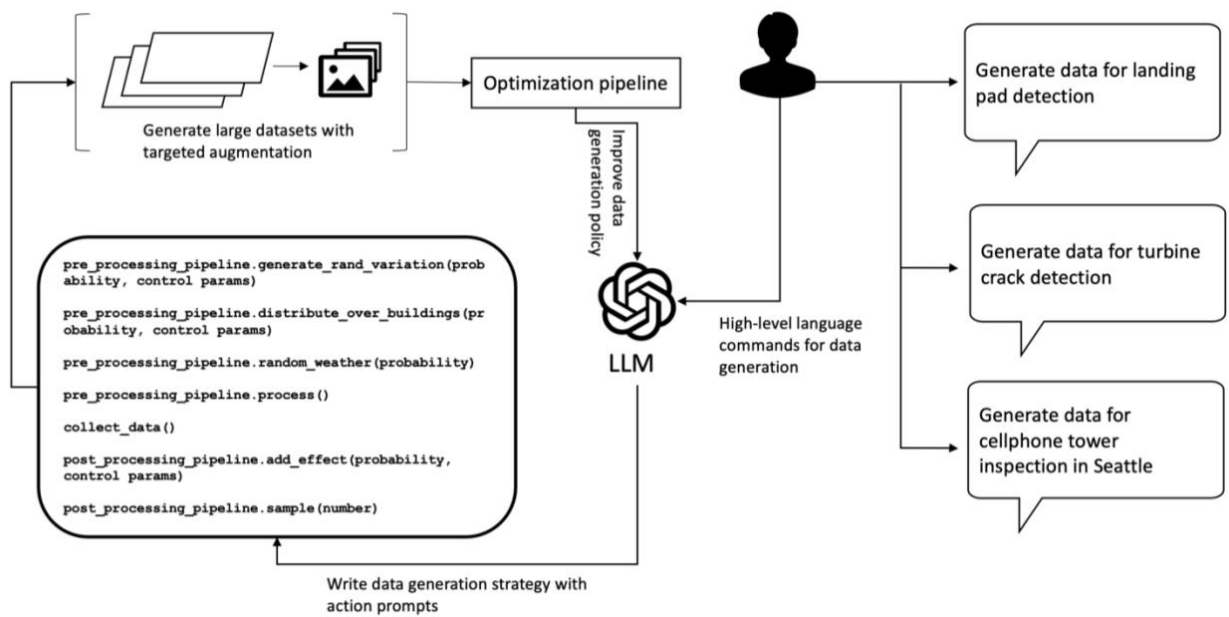

**Figure 15.** Strategy for integrating LLMs with ASDA for advanced data generation

# 6 CONCLUSION

We propose ASDA, a new scalable aerial synthetic data augmentation framework that extends a central data collection engine with the goal of increasing model generalizability through large scale scene and data augmentation. ASDA resolves limitations of existing aerial synthetic data generation workflows for generating diverse datasets through a procedural generative model. We introduced a novel layered domain randomization method curated for aerial autonomy applications that is environment type-agnostic making ASDA's scene augmentations adaptive and usable over different types of simulation environments for generating diverse synthetic datasets. We showed the potential of ASDA in increasing end data diversity and executing large-scale scene augmentations through conducted experiments. Additionally, we discussed the design benefits of our new generative procedural approach to synthetic data generation and proposed a new iterative data collection optimization that can cut the cost of data collection by learning the best operation strategies to adapt based on the downstream task. We hope that ASDA can help in advancing the generation of synthetic benchmark datasets for different navigation and detection models in drone-based applications and will help reduce overfitting to datasets with limited training or testing examples to contribute to the development of algorithms and pretraining of models that work well in real-world.

## Acknowledgements

This work was conducted as part of the first author's internship at Microsoft. The funding for this research was fully supported by Microsoft Corporation.

## References

[1] Mason Marks. 2019. Robots in space: Sharing our world with autonomous delivery vehicles. SSRN Electronic Journal (2019).

[2] Rolnick, D., Donti, P. L., Kaack, L. H., Kochanski, K., Lacoste, A., Sankaran, K., ... & Bengio, Y. (2022). Tackling climate change with machine learning. ACM Computing Surveys (CSUR), 55(2), 1-96.

[3] Watkins, S., Burry, J., Mohamed, A., Marino, M., Prudden, S., Fisher, A., ... & Clothier, R. (2020). Ten questions concerning the use of drones in urban environments. Building and Environment, 167, 106458.

[4] Nentwich, M., & Hórvath, D. M. (2018). Delivery drones from a technology assessment perspective. Overview report, 1.

[5] Bondi, E., Dey, D., Kapoor, A., Piavis, J., Shah, S., Fang, F., ... & Tambe, M. (2018, June). Airsim-w: A simulation environment for wildlife conservation with uavs. In Proceedings of the 1st ACM SIGCAS Conference on Computing and Sustainable Societies (pp. 1-12).

[6] Shah, S., Dey, D., Lovett, C., & Kapoor, A. (2018). Airsim: High-fidelity visual and physical simulation for autonomous vehicles. In Field and service robotics (pp. 621-635). Springer, Cham.

[7] Wang, W., Zhu, D., Wang, X., Hu, Y., Qiu, Y., Wang, C., ... & Scherer, S. (2020, March). Tartanair: A dataset to push the limits of visual slam. In 2020 IEEE/RSJ International Conference on Intelligent Robots and Systems (IROS) (pp. 4909-4916). IEEE.

[8] Burri, M., Nikolic, J., Gohl, P., Schneider, T., Rehder, J., Omari, S., ... & Siegwart, R. (2016). The EuRoC micro aerial vehicle datasets. The International Journal of Robotics Research, 35(10), 1157-1163.

[9] Ros, G., Sellart, L., Materzynska, J., Vazquez, D., & Lopez, A. M. (2016). The synthia dataset: A large collection of synthetic images for semantic segmentation of urban scenes. In Proceedings of the IEEE conference on computer vision and pattern recognition (pp. 3234-3243).

[10] Tassa, Y., Doron, Y., Muldal, A., Erez, T., Li, Y., Casas, D. D. L., ... & Riedmiller, M. (2018). Deepmind control suite. arXiv preprint arXiv:1801.00690.

[11] Wrennnige, M., & Unger, J. (2018). Synscapes: A photorealistic synthetic dataset for street scene parsing. arXiv preprint arXiv:1810.08705.

[12] Wu, Y., Wu, Y., Gkioxari, G., & Tian, Y. (2018). Building generalizable agents with a realistic and rich 3d environment. arXiv preprint arXiv:1801.02209.

[13] Tobin, J., Fong, R., Ray, A., Schneider, J., Zaremba, W., & Abbeel, P. (2017, September). Domain randomization for transferring deep neural networks from simulation to the real world. In 2017 IEEE/RSJ international conference on intelligent robots and systems (IROS) (pp. 23-30). IEEE.

[14] Devaranjan, J., Kar, A., & Fidler, S. (2020, August). Meta-sim2: Unsupervised learning of scene structure for synthetic data generation. In European Conference on Computer Vision (pp. 715-733). Springer, Cham.

[15] Kar, A., Prakash, A., Liu, M. Y., Cameracci, E., Yuan, J., Rusiniak, M., ... & Fidler, S. (2019). Meta-sim: Learning to generate synthetic datasets. In Proceedings of the IEEE/CVF International Conference on Computer Vision (pp. 4551-4560).

[16] Krishnan, S., Boroujerdian, B., Fu, W., Faust, A., & Reddi, V. J. (2021). Air Learning: a deep reinforcement learning gym for autonomous aerial robot visual navigation. Machine Learning, 110(9), 2501-2540.

[17] Peng, X. B., Andrychowicz, M., Zaremba, W., & Abbeel, P. (2018, May). Sim-to-real transfer of robotic control with dynamics randomization. In 2018 IEEE international conference on robotics and automation (ICRA) (pp. 3803-3810). IEEE.

[18] Prakash, A., Boochoon, S., Brophy, M., Acuna, D., Cameracci, E., State, G., ... & Birchfield, S. (2019, May). Structured domain randomization: Bridging the reality gap by context-aware synthetic data. In 2019 International Conference on Robotics and Automation (ICRA) (pp. 7249-7255). IEEE.

[19] Tan, S., Wong, K., Wang, S., Manivasagam, S., Ren, M., & Urtasun, R. (2021). Scenegen: Learning to generate realistic traffic scenes. In Proceedings of the IEEE/CVF Conference on Computer Vision and Pattern Recognition (pp. 892-901).

[20] Dosovitskiy, A., Ros, G., Codevilla, F., Lopez, A., & Koltun, V. (2017, October). CARLA: An open urban driving simulator. In Conference on robot learning (pp. 1-16). PMLR.

[21] Choi, Y., Kim, N., Hwang, S., Park, K., Yoon, J. S., An, K., & Kweon, I. S. (2018). KAIST multi-spectral day/night data set for autonomous and assisted driving. IEEE Transactions on Intelligent Transportation Systems, 19(3), 934-948.

[22] Weng, L. (2019). Domain Randomization for Sim2Real Transfer. Retrieved July 20, 2022, from https://lilianweng.github.io/posts/2019-05-05-domain-randomization/#dr-as-meta-learning

[23] Xu, Y., Huang, B., Luo, X., Bradbury, K., & Malof, J. M. (2022). SIMPL: Generating Synthetic Overhead Imagery to Address Custom Zero-Shot and Few-Shot Detection Problems. IEEE Journal of Selected Topics in Applied Earth Observations and Remote Sensing, 15, 4386-4396.

[24] Tancik, M., Casser, V., Yan, X., Pradhan, S., Mildenhall, B., Srinivasan, P. P., ... & Kretzschmar, H. (2022). Block-nerf: Scalable large scene neural view synthesis. In Proceedings of the IEEE/CVF Conference on Computer Vision and Pattern Recognition (pp. 8248-8258).

[25] Madaan, R., Gyde, N., Vemprala, S., Brown, M., Nagami, K., Taubner, T., ... & Kapoor, A. (2020, August). Airsim drone racing lab. In NeurIPS 2019 Competition and Demonstration Track (pp. 177-191). PMLR.

[26] Song, Y., Naji, S., Kaufmann, E., Loquercio, A., & Scaramuzza, D. (2021, October). Flightmare: A flexible quadrotor simulator. In *Conference on Robot Learning* (pp. 1147-1157). PMLR.

[27] Yan, Q., Zheng, J., Reding, S., Li, S., & Doytchinov, I. (2022). CrossLoc: Scalable Aerial Localization Assisted by Multimodal Synthetic Data. In Proceedings of the IEEE/CVF Conference on Computer Vision and Pattern Recognition (pp. 17358-17368).

[28] Kim, S. W., Philion, J., Torralba, A., & Fidler, S. (2021). Drivegan: Towards a controllable high-quality neural simulation. In Proceedings of the IEEE/CVF Conference on Computer Vision and Pattern Recognition (pp. 5820-5829).

[29] Ma, S., Vemprala, S., Wang, W., Gupta, J. K., Song, Y., McDuff, D., & Kapoor, A. (2022). COMPASS: Contrastive Multimodal Pretraining for Autonomous Systems. arXiv preprint arXiv:2203.15788.

[30] Ruiz, N., Schulter, S., & Chandraker, M. (2018). Learning to simulate. arXiv preprint arXiv:1810.02513.

[31] Irvin, R. B., & McKeown, D. M. (1989). Methods for exploiting the relationship between buildings and their shadows in aerial imagery. IEEE Transactions on Systems, Man, and Cybernetics, 19(6), 1564-1575.

[32] Iyer, A., Nguyen, L., & Khushu, S. (2022). Learning to identify cracks on wind turbine blade surfaces using drone-based inspection images. arXiv preprint arXiv:2207.11186.

[33] Kulsinskas, A., Durdevic, P., & Ortiz-Arroyo, D. (2021). Internal wind turbine blade inspections using UAVs: Analysis and design issues. Energies, 14(2), 294.

[34] Khandelwal, S., Suhail, M., & Sigal, L. (2021). Segmentation-grounded scene graph generation. In Proceedings of the IEEE/CVF International Conference on Computer Vision (pp. 15879-15889)

[35] Wang, R., & Qian, X. (2010). OpenSceneGraph 3.0: Beginner's guide. Packt Publishing Ltd.

[36] Unity Technologies. (2005). *Unity: Game engine* (Version 2022.1.13) [Computer software]. Unity Technologies, https://unity.com/

[37] Epic Games. (1998). *Unreal Engine: Game engine* (Version 2022.7.12) [Computer software]. Epic Games, https://unrealengine.com/

[38] Steve Coast. (2004). *OpenStreetMap: Geographic database of the world* [Computer software]. OpenStreetMap, https://openstreetmap.org/

[39] Roland Olbricht. (2008). *Overpass API: A database engine to query the OpenStreetMap data* (Version 2022-10-14) [Computer software]. OpenStreetMap, https://dev.overpass-api.de/overpass-doc/en/

[40] Marcus D. Bloice. (2016). *Augmentor: Image augmentation library* (Version 0.2.10) [Computer software] https://github.com/mdbloice/Augmentor

[41] Microsoft Corp. (2005). *Bing Maps: Web mapping service* (Version 8) [Computer software]. Microsoft Corp, https://dev.overpass-api.de/overpass-doc/en/

[42] Mahmood, R., Lucas, J., Alvarez, J. M., Fidler, S., & Law, M. T. (2022). Optimizing Data Collection for Machine Learning. arXiv preprint arXiv:2210.01234.

[43] Mikami, H., Fukumizu, K., Murai, S., Suzuki, S., Kikuchi, Y., Suzuki, T., ... & Hayashi, K. (2021). A Scaling Law for Synthetic-to-Real Transfer: How Much Is Your Pre-training Effective?. arXiv preprint arXiv:2108.11018.

[44] Cubuk, E. D., Zoph, B., Mane, D., Vasudevan, V., & Le, Q. V. (2019). Autoaugment: Learning augmentation strategies from data. In Proceedings of the IEEE/CVF Conference on Computer Vision and Pattern Recognition (pp. 113-123).

[45] OpenAI. (2022). *ChatGPT* [Artificial-Intelligence (AI) chatbot]. OpenAI, https://openai.com/blog/chatgpt/